\def\year{2022}\relax
\documentclass[letterpaper]{article} % DO NOT CHANGE THIS
\usepackage{aaai22}  % DO NOT CHANGE THIS
\usepackage{times}  % DO NOT CHANGE THIS
\usepackage{helvet}  % DO NOT CHANGE THIS
\usepackage{courier}  % DO NOT CHANGE THIS
\usepackage[hyphens]{url}  % DO NOT CHANGE THIS
\usepackage{graphicx} % DO NOT CHANGE THIS
\usepackage{natbib}  % DO NOT CHANGE THIS AND DO NOT ADD ANY OPTIONS TO IT
\usepackage{caption} % DO NOT CHANGE THIS AND DO NOT ADD ANY OPTIONS TO IT
\DeclareCaptionStyle{ruled}{labelfont=normalfont,labelsep=colon,strut=off} % DO NOT CHANGE THIS
\usepackage{algorithm}
\usepackage{algorithmic}
\usepackage{amssymb,amsmath,epsfig}
\usepackage{booktabs}
\usepackage{mathrsfs}
\usepackage{color}
\usepackage{textcomp}
\usepackage{bbding}
\usepackage{pifont}
\usepackage{wasysym}
\usepackage{amssymb}
\usepackage{subcaption}
\usepackage{cite}
\usepackage{amsfonts}
\usepackage{xcolor}
\usepackage{multirow}

\usepackage{newfloat}
\usepackage{listings}
\floatstyle{ruled}
\newfloat{listing}{tb}{lst}{}
\floatname{listing}{Listing}
\title{Feature Distillation Interaction Weighting Network for \\ Lightweight Image Super-Resolution}
\author {
    Guangwei Gao\textsuperscript{\rm 1}\textsuperscript{$\dagger$},
    Wenjie Li\textsuperscript{\rm 1}\textsuperscript{$\dagger$},
    Juncheng Li\textsuperscript{\rm 2}\thanks{Corresponding author,   $\dagger$Equal contributions.},
    Fei Wu\textsuperscript{\rm 1},
    Huimin Lu\textsuperscript{\rm 3},
    Yi Yu\textsuperscript{\rm 4}
}

\affiliations {
    \textsuperscript{\rm 1} Nanjing University of Posts and Telecommunications\\ 
    \textsuperscript{\rm 2} The Chinese University of Hong Kong\\
    \textsuperscript{\rm 3} Kyushu Institute of Technology\\
    \textsuperscript{\rm 4} National Institute of Informatics \\
    \{csggao,cvjunchengli\}@gmail.com,
    liwj0824@163.com,
    wufei\_8888@126.com,
    luhuimin@ieee.org,
    yiyu@nii.ac.jp
}

\usepackage{bibentry}
\begin{document}

\maketitle

\begin{abstract}
Convolutional neural networks based single-image super-resolution (SISR) has made great progress in recent years. However, it is difficult to apply these methods to real-world scenarios due to the computational and memory cost. Meanwhile, how to take full advantage of the intermediate features under the constraints of limited parameters and calculations is also a huge challenge. To alleviate these issues, we propose a lightweight yet efficient Feature Distillation Interaction Weighted Network (FDIWN). Specifically, FDIWN utilizes a series of specially designed Feature Shuffle Weighted Groups (FSWG) as the backbone, and several novel mutual Wide-residual Distillation Interaction Blocks (WDIB) form an FSWG. In addition, Wide Identical Residual Weighting (WIRW) units and Wide Convolutional Residual Weighting (WCRW) units are introduced into WDIB for better feature distillation. Moreover, a Wide-Residual Distillation Connection (WRDC) framework and a Self-Calibration Fusion (SCF) unit are proposed to interact features with different scales more flexibly and efficiently.
Extensive experiments show that our FDIWN is superior to other models to strike a good balance between model performance and efficiency. The code is available at \url{https://github.com/IVIPLab/FDIWN}.
\end{abstract}

\section{Introduction}
\label{sec1}

Due to the huge computational overhead of traditional super-resolution, it is difficult to be applied to mobile devices with limited computing capabilities. The main goal of lightweight single-image super-resolution (SISR) is to reconstruct super-resolution (SR) images from the low-resolution (LR) one with fewer parameters and calculations~\cite{yao2020weighted,li2020s,hou2021coordinate}. In the past ten years, deep learning has made amazing achievements in various computer vision tasks, which also greatly promoted the development of SISR. 

Recently, many convolutional neural networks (CNNs) based SISR methods have been proposed~\cite{3dong2015image,5han2015deep,4dong2016accelerating,tian2020coarse,he2019mrfn}. Compared with the traditional methods, CNN-based SISR methods are more versatile and can reconstruct higher-quality SR images with more texture details. In 2014, Dong \emph{et al.} introduced the deep learning technology into SISR and proposed the first CNN-based SISR model, named SRCNN~\cite{3dong2015image}. Although SRCNN has only three convolutional layers, its performance has far surpassed traditional methods and achieved state-of-the-art results at the time. Now, we know that deeper and more complex networks can achieve better performance~\cite{6lim2017enhanced,1ahn2018fast,7haris2018deep,12zhang2018image,17zhang2018residual,li2018multi,zhang2020unsupervised,li2020mdcn}. However, their parameters and calculations are also huge and are difficult to be used on mobile devices. To address this issue, many lightweight SISR models have also been proposed. For instance, CARN~\cite{1ahn2018fast} is a lightweight residual network composed of multiple residual connections. ECBSR~\cite{zhang2021edge} is a lightweight and efficient network whose features are extracted in multiple paths. The purpose of these models is to reduce the complexity of the model and facilitate the application in the real world. The demand for lightweight practical models motivates us to propose the Feature Distillation Interaction Weighted Network (FDIWN). The computational costs of FDIWN are lower than most existing lightweight SISR models, but it is not inferior to them in terms of performance.

\begin{figure*}[t]
	\centerline{\includegraphics[width=15.5cm]{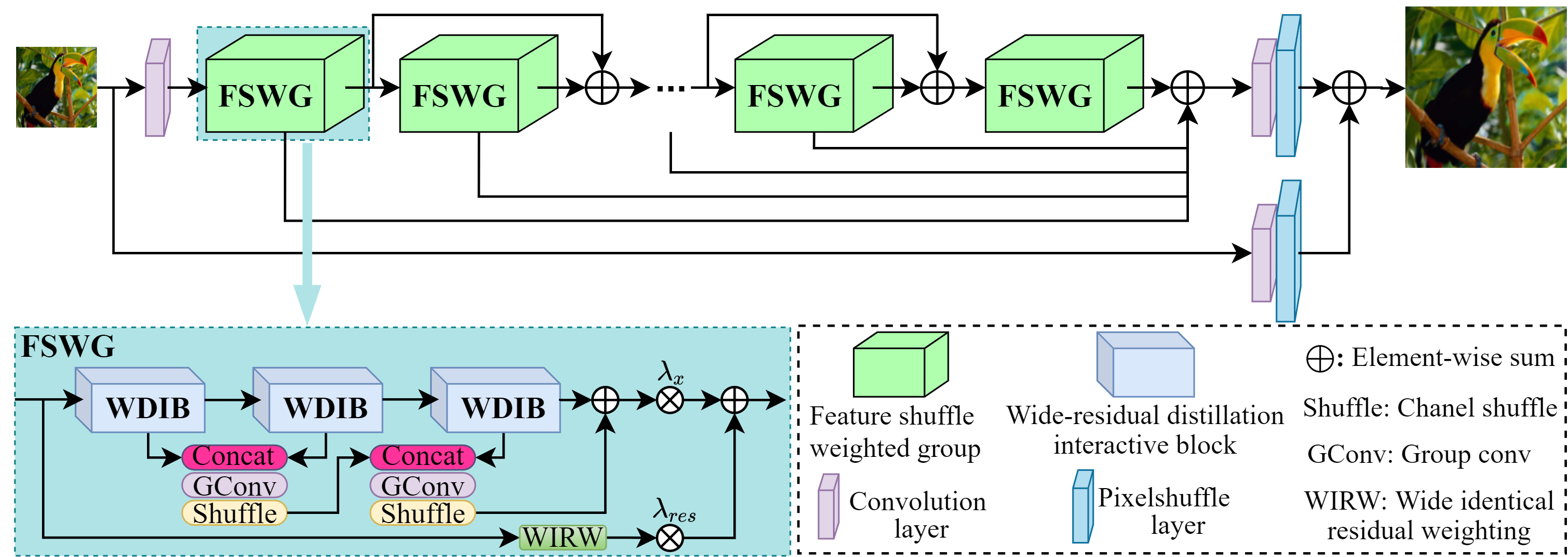}}
	\caption{The architecture of the proposed Feature Distillation Interaction Weighting Network (FDIWN).}
	\label{Figure 2}
\end{figure*}

As we know, as the depth of the network increases, information will be lost during transmission. Therefore, under the constraints of parameters and calculations, how to prevent information loss, and how to make full use of intermediate features is important. To achieve this, we introduce Wide-residual Distillation Interaction Blocks (WDIB) in the Feature Shuffle Weighted Group (FSWG) for pairwise feature fusion, and then the features are shuffled and weighted. This operation can improve model performance while only increasing a small amount of computational cost since the wide-residual attention weighting units are extensively used in WDIB, including Wide Identical Residual Weighting (WIRW) units and Wide Convolutional Residual Weighting (WCRW) units. WIRW and WCRN allow more features to pass and be activated, thereby increasing the transmission and utilization of the features. Meanwhile, the carefully designed Self-Calibration Fusion (SCF) unit integrates different levels of features by the jump splicing strategy to achieve a good SR reconstruction. In general, our main contributions can be summarized as follows: 

\begin{itemize}
\item We propose a wide-residual attention weighting unit for lightweight SISR, including Wide Identical Residual Weighting (WIRW) unit and Wide Convolutional Residual Weighting (WCRW) unit, which has stronger feature distillation capabilities than ordinary residual blocks.
\item We propose a novel Self-Calibration Fusion (SCF) module to replace the traditional concatenate operation for efficient feature interaction and fusion, which can aggregate more representative features and self-calibrate the input and output features.
\item We propose a Wide-Residual Distillation Connection (WRDC) framework, which connects the coarse and distilled fine features within the module and allows features from different scales to interact with each other. 
\item We design a Feature Shuffle Weighted Group (FSWG) for pairwise feature fusion, which consists of a series of interactional WDIBs. Meanwhile, it serves as a basic component of the proposed Feature Distillation Interaction Weighting Network (FDIWN).
\end{itemize}

\begin{figure*}[t]
	\centerline{\includegraphics[width=15.5cm]{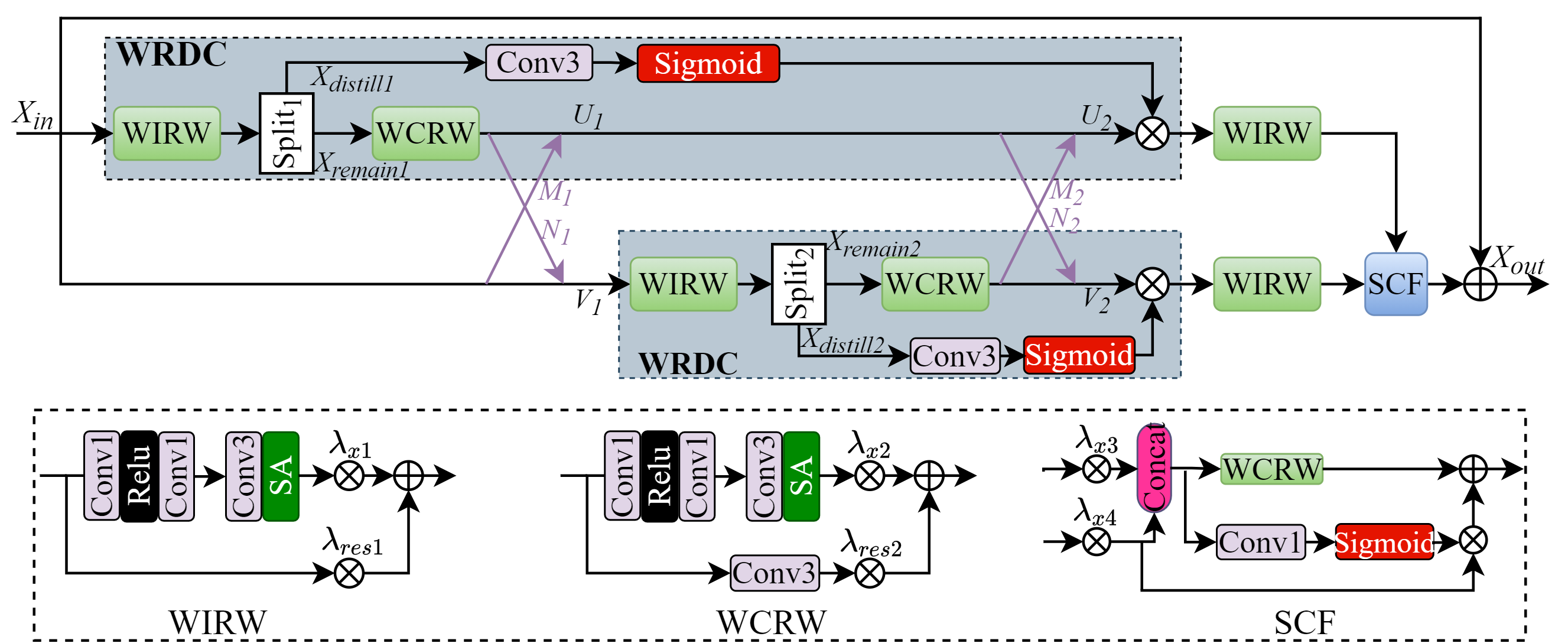}}
	\caption{The structure of the proposed Wide-residual Distillation Interaction Block (WDIB). Conv1 and Conv3 represent the convolutional layer with the kernel size of 1 and 3, respectively.}
	\label{Figure 3}
\end{figure*}

\section{Related Work}
\label{sec2}

\subsection{Lightweight SISR}
\label{sec21}

Recently, more and more effective deep neural networks have been introduced into SISR. However, most of them are often accompanied by a large number of model parameters and need large calculation costs, which limits their applications on mobile devices. To address this issue, researchers began to explore lightweight and efficient SISR methods. For instance, CARN-M~\cite{1ahn2018fast} used group convolution to reduce the model parameters, which even achieved better super-resolved effects than some large SISR models. Hui \emph{et al.}~\cite{37hui2018fast} presented the Information Distillation Network (IDN) to extract more useful information with fewer convolutional layers. Meanwhile, IMDN~\cite{11hui2019lightweight} was modified based on the IDN with a faster and lighter structure. After that, RFDN~\cite{19liu2020residual} changed the channel splitting method based on IMDN and adopted skip connections for the convolutional layers in the residual block. Wang \emph{et al.}~\cite{38wang2019lightweight} proposed an adaptive weighted super-resolution network with efficient residual learning and local residual fusion. Wang~\cite{wang2021lightweight} proposed a multi-scale feature interaction network (MSFIN) for lightweight SISR. IMRN~\cite{jiang2021learning} uses the pruning method to reduce the model size without significantly reducing the performance, and achieves good performance. Nowadays, lightweight SISR is getting more and more important since its great application value. Although the aforementioned models achieved good results, they ignored the use of intermediate features, resulting in sub-optimal performance.

\subsection{Wide-Residual Attention Weighting Learning}
\label{sec22}

Studies have shown that the deeper the network, the better the performance of the model. However, it was later discovered that as the number of the network layers increased, the performance of the model does not rise but falls. To solve this problem, the residual block was introduced into the network, thus the network can reach very deep, and the effect of the network will also become better. This method is also used in many SISR models. For example, VDSR~\cite{25kim2016accurate} is a 20 layers network, EDSR~\cite{6lim2017enhanced} is a 65 layers network, and RCAN~\cite{12zhang2018image} has more than 800 layers. All these models introduced various skip connections and concatenation operations between shallow layers and deep layers to make full use of the shallow feature information. Different from the above methods, Yu \emph{et al.}~\cite{27yu2018wide} found that the model with wider features before ReLU activation can achieve better performance. Therefore, they proposed the WDSR with the wide activation mechanism, which expanded features before ReLU and allowed more information to pass through without additional parameters. Meanwhile, the attention mechanism has been widely used in deep learning tasks. For instance, Zhang \emph{et al.}~\cite{12zhang2018image} and Dai \emph{et al.}~\cite{14dai2019second} proposed the first-order statistics and second-order attention networks to pursue better feature extraction. Inspired by this, we try to introduce the second-order attention mechanism into our modified wide activation residual block to further improve the feature extraction ability of the model.

\section{Proposed Method}
\label{sec3}
To build a lightweight and accurate SISR model, we focus on the use of intermediate features and the interaction of feature information. To achieve this, we propose a Feature Distillation Interaction Weighting Network (FDIWN). FDIWN consists of a series of novel and efficient modules, including Wide-residual Distillation Interaction Block (WDIB), Wide-Residual Distillation Connection (WRDC), Wide Identical Residual Weighting (WIRW) unit, Wide Convolutional Residual Weighting (WCRW) unit, and Self-Calibration Fusion (SCF) module.

\subsection{Feature Distillation Interaction Weighting Network}
\label{sec31}

As shown in Figure~\ref{Figure 2}, FDIWN consists of three parts: the shallow feature extraction part, the non-linear deep feature acquisition part, and the upsampling recovery part. Following previous works, we use a $3 \times 3$ convolutional layer to extract the shallow features ${X_0}$ from the input LR image
\begin{equation}
{X_0} = {C_e}({I_{LR}}),
\end{equation}
where ${C_e}$ represents the feature extraction layer, ${I_{LR}}$ is the LR image, and ${X_0}$ is the extracted shallow features.

After that, the non-linear feature mapping module is followed, which is formed by several Feature Shuffle Weighted Groups (FSWGs) through jump connections. The operation can be expressed as follow
\begin{equation}
{X_1} = F_{FSWG}^0({X_0}),
\end{equation}
\begin{equation}
{X_{n - 1}} = F_{FSWG}^{n - 2}( \ldots (F_{FSWG}^1({X_1}) + {X_1}) \ldots ) + {X_{n - 2}},
\end{equation}
\begin{equation}
{X_n} = F_{FSWG}^{n - 1}({X_{n - 1}}),
\end{equation}
where $F_{FSWG}^k$ represents the $k$-th FSWG, and ${X_n}$ denotes the extracted non-linear deep features.

The features used for the final SR image reconstruction in the upsampling recovery module come from two parts, one comes from the non-linear deep feature extraction module and the other comes from the input LR image. We hope that superimposing the low-frequency and high-frequency feature information in this way can reconstruct high-quality SR images with more texture details. Therefore, the final SR image can be expressed as
\begin{equation}
{I_{SR}} = {F_{UP1}}(\sum\limits_{i = 0}^{n - 1} {F_{FSWG}^i({X_i})} )  + {F_{UP2}}({I_{LR}}),
\end{equation}
where ${I_{SR}}$ is the reconstructed SR image, ${F_{UP1}}$ and ${F_{UP2}}$ represent the upsampling modules.

\subsection{Wide-Residual Distillation Connection}
\label{sec32}
\begin{figure}[t]
	\centerline{\includegraphics[width=8.2cm,trim=0 0 0 0]{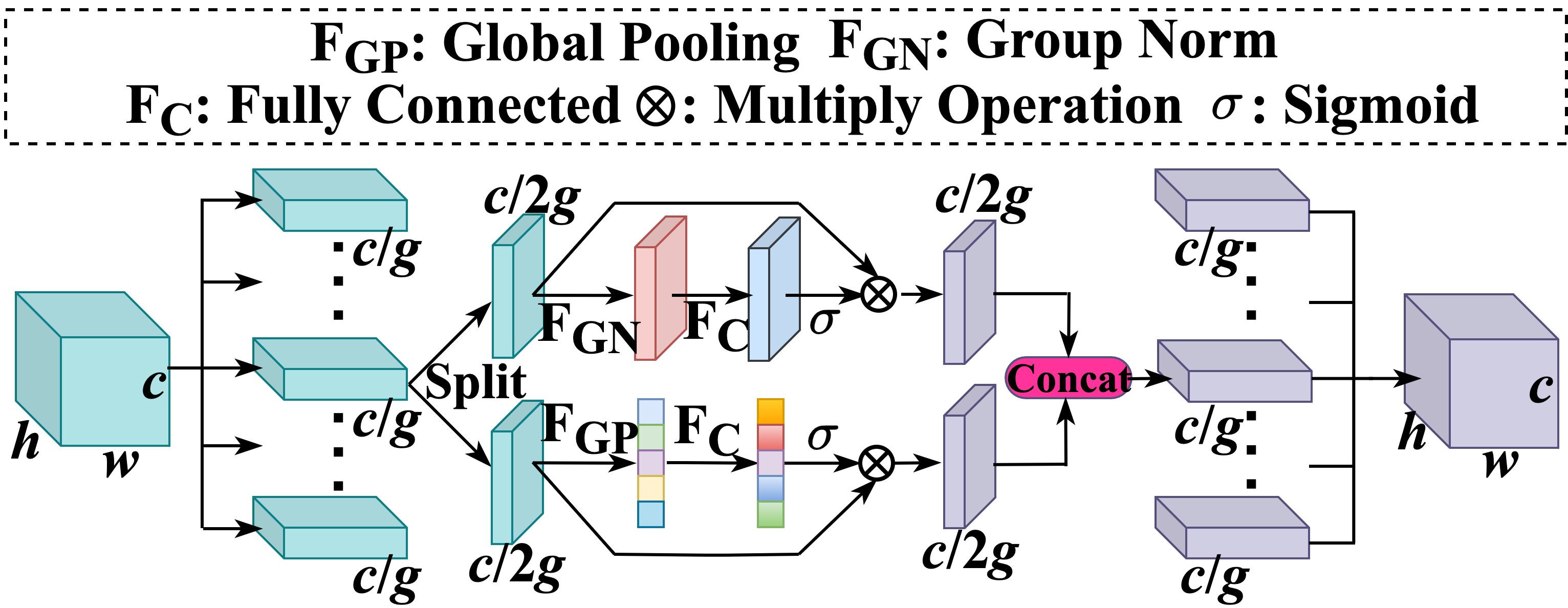}}
	\caption{The principle of shuffle attention (SA) mechanism. $h$, $w$, $c$, and $g$ represent the height, width, number of channels, and number of groups, respectively.}
	\label{Figure 5}
\end{figure}

As shown in Figure~\ref{Figure 3}, Wide-Residual Distillation Connection (WRDC) is an important component in the model, which consists of the Wide Identical Residual Weighting (WIRW) unit, the Wide Convolutional Residual Weighting (WCRW) unit, and the distillation jump connection. Both WIRW and WCRW introduced the wide activation mechanism, thus it can distill richer features with fewer parameters. Recently, with the emphasis on the importance of channel attention in RCAN~\cite{12zhang2018image}, many SR methods focus on the attention mechanism. As shown in Figure~\ref{Figure 5}, Zhang et al.~\cite{31zhang2021sa} proposed a new attention paradigm, which is a combination of spatial attention and channel attention, called Shuffle Attention (SA). Inspired by this, we introduce the SA into our WIRW and WCRW to further enhance their feature extraction abilities. Since the SA mechanism is placed in each wide-residual unit, we set the numbers of the group $g$ to be large enough to keep the SA lightweight. After the channel splitting operation, the number of input channels of WCRW is only half of the original input. Therefore, compared with WIRW, a $3 \times 3$ convolutional layer is added to the shortcut path of WCRW to increase the number of output channels so that it can match the original input size and achieve the interaction between different features. Meanwhile, WCRW and WIRW both introduce adaptive weights operation in the main path and shortcut path for adaptive feature learning. When input $I$ is fed into WIRW and WCRW, the output ${X_{WIRW}}$ and ${X_{WCRW}}$ can be expressed as
\begin{equation}
{X_{WIRW}} = {\lambda _{x1}}{F_{SA}}[{F_{CR}}(I)] + {\lambda _{res1}}I,
\end{equation}
\begin{equation}
{X_{WCRW}} = {\lambda _{x2}}{F_{SA}}[{F_{CR}}(I)] + {\lambda _{res2}}{F_{C3}}(I),
\end{equation}
where ${\lambda _{xk}}$ and ${\lambda _{resk}}$ ($k = 1,2$) represent the adaptive multiplier of the $k$-th wide-residual unit branch, ${F_{SA}}$ represents the SA operation, ${F_{CR}}$ represents a series of (conv + relu) operations before the attention mechanism, and ${F_{C3}}$ represents the $3 \times 3$ convolutional layer. The complete structure of WIRW and WCRW can be seen in Figure~\ref{Figure 3}. 

Apart from the above operation, the distillation connection part is applied to segment the channel features through the convolutional layer and Sigmoid function. The convolutional layer is introduced to expand the dimension of the splitting channel, while the Sigmoid function non-linearizes the obtained coarse high-frequency features to obtain fine features maps. Finally, these features are multiplied with the low-frequency attention features obtained after the wide-residual unit refinement process to realize the interaction of the features from different scales.

\subsection{Wide-Residual Distillation Interaction Block}
\label{sec33}

\begin{figure}[t]
	\centerline{\includegraphics[width=8.2cm,trim=0 0 0 0]{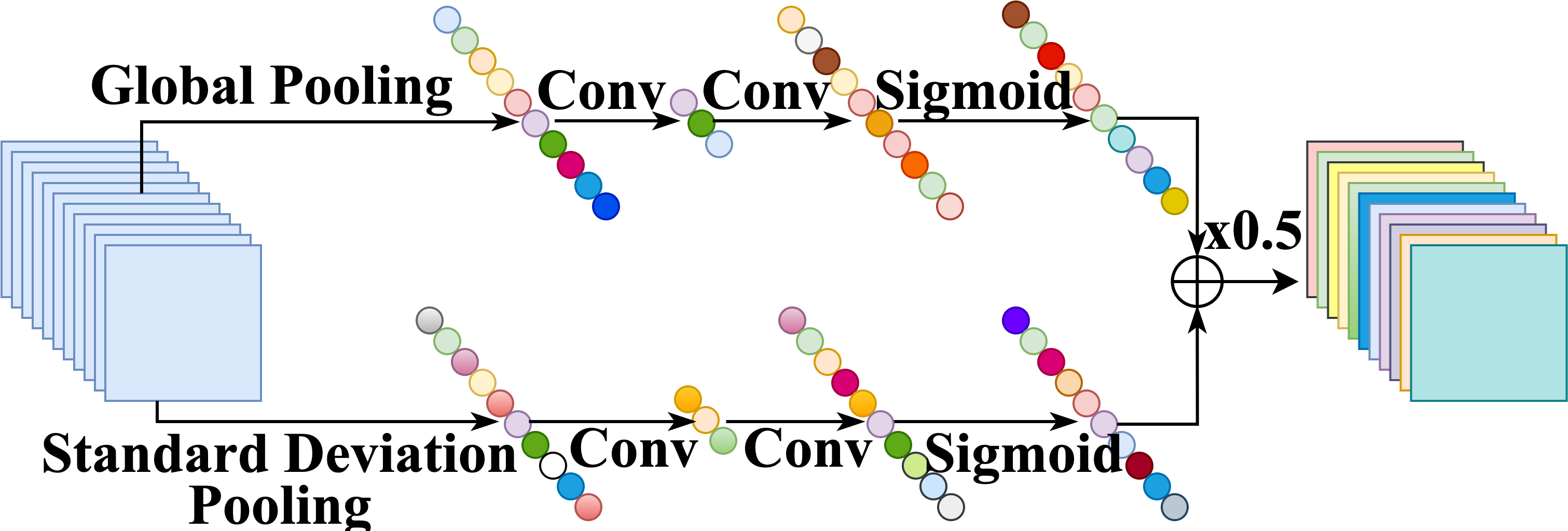}}
	\caption{The diagram of combination coefficient learning.}
	\label{Figure 4}
\end{figure}

Inspired by the lattice block (LB)~\cite{32luo2020latticenet}, we design a Wide-Residual Distillation Interaction Block (WDIB) based on WRDC. As shown in Figure~\ref{Figure 3}, WDIB utilizes the butterfly structure described in LB to realize the interaction of intermediate features. Define ${W_{ir}}$ and ${W_{cr}}$ represent the WIRW and WCRW units, the first butterfly structure can be expressed as
\begin{equation}
{X_{remain1}},{X_{distill1}} = Spli{t_1}({W_{ir}}({X_{in}})),
\end{equation}
\begin{equation}
{U_{1}} = {M_1}\left\langle {{X_{in}}} \right\rangle  + {W_{cr}}({X_{in}}),
\end{equation}
\begin{equation}
{V_{1}} = {N_1}\left\langle {{W_{cr}}({X_{remain1}})} \right\rangle  + {X_{in}},
\end{equation}
where $Spli{t_i}(\cdot)$ represents the $i$-th channel splitting operation, ${X_{remaini}}$ represents the rough feature of the input subsequent wide-residual unit, and ${X_{distilli}}$ represents the $i$-th refinement feature that is split out and jump-connected with the next butterfly structure. The combination coefficients ${M_{i}}$ and ${N_{i}}$ are the two vectors connecting the upper and lower branches. It is worth noting that $M\left\langle {X_{in}} \right\rangle = M({X_{in}}) \times {X_{in}}$, and the learning details of the combination coefficients ${M_i}$ and ${N_i}$ are provided in Figure~\ref{Figure 4}. ${U_{i}}$ and ${V_{i}}$ are the output features after the $i$-th butterfly structure, and then they are fed into the second butterfly structure
\begin{equation}
{X_{remain2}},{X_{distill2}} = Spli{t_2}({W_{ir}}({V_{1}})),
\end{equation}
\begin{equation}
{U_2} = {M_2}\left\langle {{W_{cr}}({X_{remain2}})} \right\rangle  + {U_{1}},
\end{equation}
\begin{equation}
{V_2} = {N_2}\left\langle {{U_1}} \right\rangle  + {W_{cr}}({X_{remain2}}).
\end{equation}

After that, the output $X_{out}$ can be expressed as
\begin{equation}
{X_{out1}} = {W_{ir}}({U_2} \times {S_3}({X_{distill1}})),
\end{equation}
\begin{equation}
{X_{out2}} = {W_{ir}}({V_2} \times {S_3}({X_{distill2}})),
\end{equation}
\begin{equation}
{X_{out}} = {C_{SCF}}[{X_{out1}},{X_{out2}}] + {X_{in}},
\end{equation}
where ${C_{SCF}}$ represents the proposed Self-Calibration Fusion (SCF) module. ${X_{out1}}$ and ${X_{out2}}$ represent the upper branch and the lower branch entering the SCF module, respectively. In addition, ${S_k}$ represents a $k \times k$ convolutional layer followed by a cascade of Sigmoid function. The structure of the SCF module is provided in Figure~\ref{Figure 3}. Obviously, the features of the upper and lower branches are first fused, and then the different scale features from two branches are interacted and fused. The output ${X_{SCF}}$ of the SCF module can be expressed as
\begin{equation}
{X_{concat}} = {C_{Concat}}[{\lambda _{x3}}{X_{out1}},{\lambda _{x4}}{X_{out2}}],
\end{equation}
\begin{equation}
{X_{SCF}} = {\lambda _{x4}}{X_{out2}} \times {S_1}({X_{concat}}) + {W_{cr}}({X_{concat}}),
\end{equation}
where ${C_{Concat}}$ represents the Concat operation, ${\lambda _{x3}}$ and ${\lambda _{x4}}$ represent the adaptive weights, and ${X_{concat}}$ represents the output after the upper and lower branches are multiplied by the adaptive weight and then the Concat is performed. Subsequently, the fused features are nonlinearized, then multiplied with the features of the lower branch, and finally added to the refined fusion features to achieve the interaction of features from different scales. Since there are a large number of adaptive multipliers in the module, the output features can be adjusted and calibrated continuously during the training, thus it can achieve better performance than the traditional Concat operation.

\subsection{Feature Shuffle Weighted Group}
\label{sec34}

As shown in Figure~\ref{Figure 2}, Feature Shuffle Weighted Group (FSWG) consists of three interactional WDIBs and it serves as the basic component of FDIWN. Specifically, we fuse and shuffle the features extracted by WDIB one by one. The cascaded operation ${F_{CGS}}$ can be expressed as
\begin{equation}
{F_{CGS}} = {F_{Shuffle}}({F_{GConv}}({C_{Concat}}[{x_i},{x_{i + 1}}])),
\end{equation}
where ${F_{Shuffle}}$ represents the channel shuffle operation and ${F_{GConv}}$ represents the group convolution operation. ${x_i}$ and ${x_{i + 1}}$ represent the two features to be merged. After that, we add the shuffled and fused features with the original features that have not been operated to achieve the interaction of feature information. Meanwhile, we set a larger group in the shuffle and fusion operation to reduce the parameter burden. Moreover, to reduce the redundant information, the primary features and the features after the information interaction are self-adaptively fused to distill the desired important features. Define the input of FSWG as ${W_0}$, the output ${W_{out}}$ can be formulated as 
\begin{equation}
{W_{CGS}} = {F^2}_{CGS}({F^1}_{CGS}({W_1},{W_2}),{W_3}),
\end{equation}
\begin{equation}
{W_{out}} = {\lambda _x}({W_{CGS}} + {W_3}) + {\lambda _{res}}{W_{ir}}({W_0}),
\end{equation}
where ${W_i}$ represents the output of the $i$-th WDIB, ${F^i}_{CGS}$ represents the $i$-th ${F_{CGS}}$ operation, ${W_{CGS}}$ represents the extracted features after a series of shuffle and fusion operations. In addition, ${\lambda _x}$ and ${\lambda _{res}}$ are used to adaptively adjust the weight of each channel. After these operations, the features from each WDIB are adequately interacted and distilled to achieve better SR image reconstruction.

\section{Experiments}
\label{sec4}

\subsection{Datasets and Evaluation Metrics}
\label{sec41}
Following previous works, we use the DIV2K~\cite{agustsson2017ntire} as the training dataset, which contains 800 pairs of images. For testing, we use Set5~\cite{bevilacqua2012low}, Set14~\cite{zeyde2010single}, BSDS100~\cite{martin2001database}, and Urban100~\cite{huang2015single} to verify the effectiveness of the proposed FDIWN. Meanwhile, two metrics on the Y channel in the YCbCr color space, namely PSNR and SSIM are used to evaluate the model performance.

\subsection{Implementation Details}
\label{sec42}

Each mini-batch during the training consists of 16 RGB image blocks with the size of $48 \times 48$, which are randomly cropped from the LR image. Meanwhile, the training dataset is enhanced by random and horizontal rotation at different angles for data augmentation. The learning rate is initialized to 2e-4 and a total of 1000 epochs are updated. We implement our model with the PyTorch framework and update it with Adam optimizer. All our experiments are performed on NVIDIA RX 2080TI GPUs. 

As for the model set, the final version of FDIWN consists of 6 FSWGs, while the tiny version of FDIWN-M only consists of 4 FSWGs. The number of input channels is initialized to 24 and the value of the adaptive weight is 1. 

\begin{table}[t]
 \centering
  \setlength\tabcolsep{3pt}
  \renewcommand\arraystretch{0.4}
  {\fontsize{9pt}{12pt}\selectfont
  \begin{tabular}{@{}ccccccc@{}}
   \toprule
    Method  &WR&DC  &SCF &Params  &Multi-adds   & PSNR~~~SSIM        \\
   \hline
   Baseline1       &\XSolid  &\XSolid   &\XSolid          &59K    &3.3G    & 37.52~~~0.9587  \\
            Baseline2   &\Checkmark     &\XSolid    &\XSolid            &59K    &4.9G    & 37.53~~~0.9589   \\
   FDIWN        &\XSolid  &\XSolid     &\Checkmark       &89K    &6.5G     & 37.58~~~0.9591     \\
   FDIWN       &\Checkmark   &\Checkmark     &\XSolid      &65K    &6.5G    & 37.59~~~0.9590       \\
   FDIWN       &\Checkmark   &\Checkmark  &\Checkmark        &96K    &9.7G    & \textbf{37.64}~~~\textbf{0.9592}  \\ \bottomrule 
  \end{tabular}}
  \caption{Impact analysis of WRDC and SCF.}
  \label{tab1}
\end{table}

\begin{table}[t]
\centering
  \setlength\tabcolsep{3pt}
  {\fontsize{9pt}{12pt}\selectfont
  \begin{tabular}{@{}ccccccc@{}}
   \toprule
   Case & Method &Channels &Params   &Multi-adds     & PSNR~~~SSIM   \\ \hline
   1    & Baseline  & 24    & 152K  & 23.2G     & 37.70~~~0.9594          \\
   2    & FDIWN     & 48  & 96K  & 9.7G     & 37.64~~~0.9592           \\
   3    & FDIWN     &120    &131K  & 9.7G & \textbf{37.72}~~~\textbf{0.9596}  \\ \bottomrule 
  \end{tabular}}
  \caption{Impact analysis of WIRW and WCRW.}
  \label{tab2}
\end{table}

\subsection{Ablation Studies}
\label{sec43}

To verify the efficiency and effectiveness of the proposed modules, we conducted a series of ablation studies and all of these studies are tested on the Set5 dataset.

\textbf{The effectiveness of WRDC and SCF.} To verify the effectiveness of WRDC and SCF, we replace the WRDC module in FDIWN with a three-layer cascaded $3 \times 3$ convolution plus ReLU layer, and replace the SCF module with the concatenate operation. We set this structure as the Baseline1. Baseline2 has a similar structure while the three-layer cascaded convolution plus ReLU layer is placed by the cascaded WIRW plus WCRW units. Then we add WRDC and SCF step by step and compare their performance with the Baseline1. It can be observed from Table~\ref{tab1} that our proposed FDIWN improves the performance of Baseline1 by 0.12 dB, which proves the effectiveness of the WRDC and SCF module. Specifically, with the DC mechanism, PSNR is increased from 37.53 dB to 37.59 dB, while the number of parameters only increases 6K. Meanwhile, the SCF module can provide a 0.06 dB improvement in model performance with an acceptable increase in the number of parameters.

\textbf{The effectiveness of WIRW and WCRW.} To evaluate the role of our wide-residual units in the module, we replaced all WIRW and WCRW units in our module with the basic residual block and treat this structure as the baseline model. The kernel size of the convolutional layer in the basic residual block is $3 \times 3$. To explore the impact of the number of channels before the activation function in our designed wide-residual units on the SR performance, we set the number of channels as 48 and 120, respectively. Due to the lightweight character of the 1$\times$1 convolution function, we can see from Table~\ref{tab2} that \romannumeral 1) Compared with the Baseline model (Case 1 and 3), FDIWN achieves better performance with fewer parameters and computational costs; \romannumeral 2) Increasing the number of channels (Case 2 and 3), the model performance can be further improved. This fully demonstrates the effectiveness of the introduced wide activation mechanism and the proposed WIRW and WCRW.

\begin{figure}[t]
	\centerline{\includegraphics[width=8cm,trim=0 0 0 0]{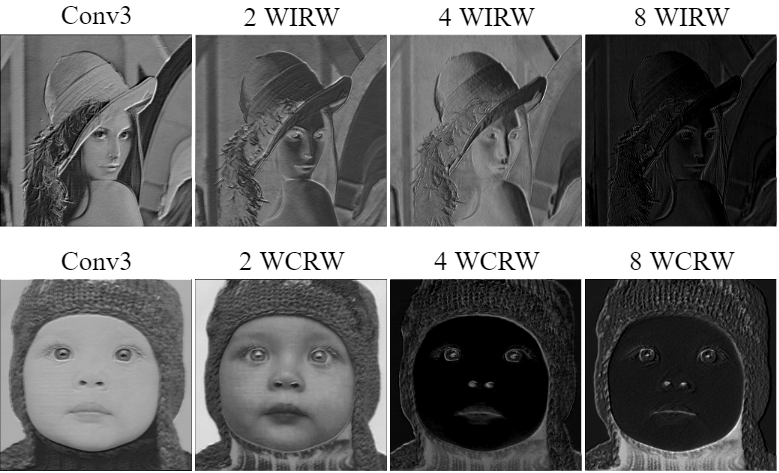}}
	\caption{Feature visualization of different module.}
	\label{keshihua}
\end{figure}

To further verify the impact of WIRW and WCRW units on SISR, we visualize the features produced by the different numbers of WIRW and WCRW units. The network here is composed of one $3 \times 3$ convolution and several wide-residual units. Figure~\ref{keshihua} shows that the $3 \times 3$ convolution can only extract shallow image information but cannot extract image details well. However, it can be seen that as the number of WIRW and WCRW units increases, more edge detail information will be extracted. This also means that as the number of WIRW and WCRW units increases, the ability of the model to capture high-frequency information will be greatly improved. This further demonstrates the effectiveness of WIRW and WCRW.

\begin{table}[t]
\centering
  \setlength\tabcolsep{4pt}
  {\fontsize{9pt}{12pt}\selectfont
  \begin{tabular}{@{}cccccc@{}}
   \toprule
    Method     &BI   &WIRW   &Params     &Multi-adds     & PSNR~~~SSIM        \\ %\midrule
   \hline
                                 \hline
   Baseline    &\XSolid    &\XSolid  &215K    &22.0G                 & 37.81~~~0.9598          \\
   FDIWN       &\Checkmark   &\XSolid  &225K    &24.4G               & 37.85~~~\textbf{0.9600} \\
   FDIWN       &\Checkmark   &\Checkmark     &230K        &24.4G               & \textbf{37.88}~~~\textbf{0.9600}  \\ \bottomrule 
   %\bottomrule 
  \end{tabular}}
  \caption{Evaluation of the combination structure of WDIB. The best results are highlighted. BI: Block Interaction.}
  \label{tab3}
\end{table}

\begin{figure}[t]
   \centering
   \includegraphics[width=8.3cm]{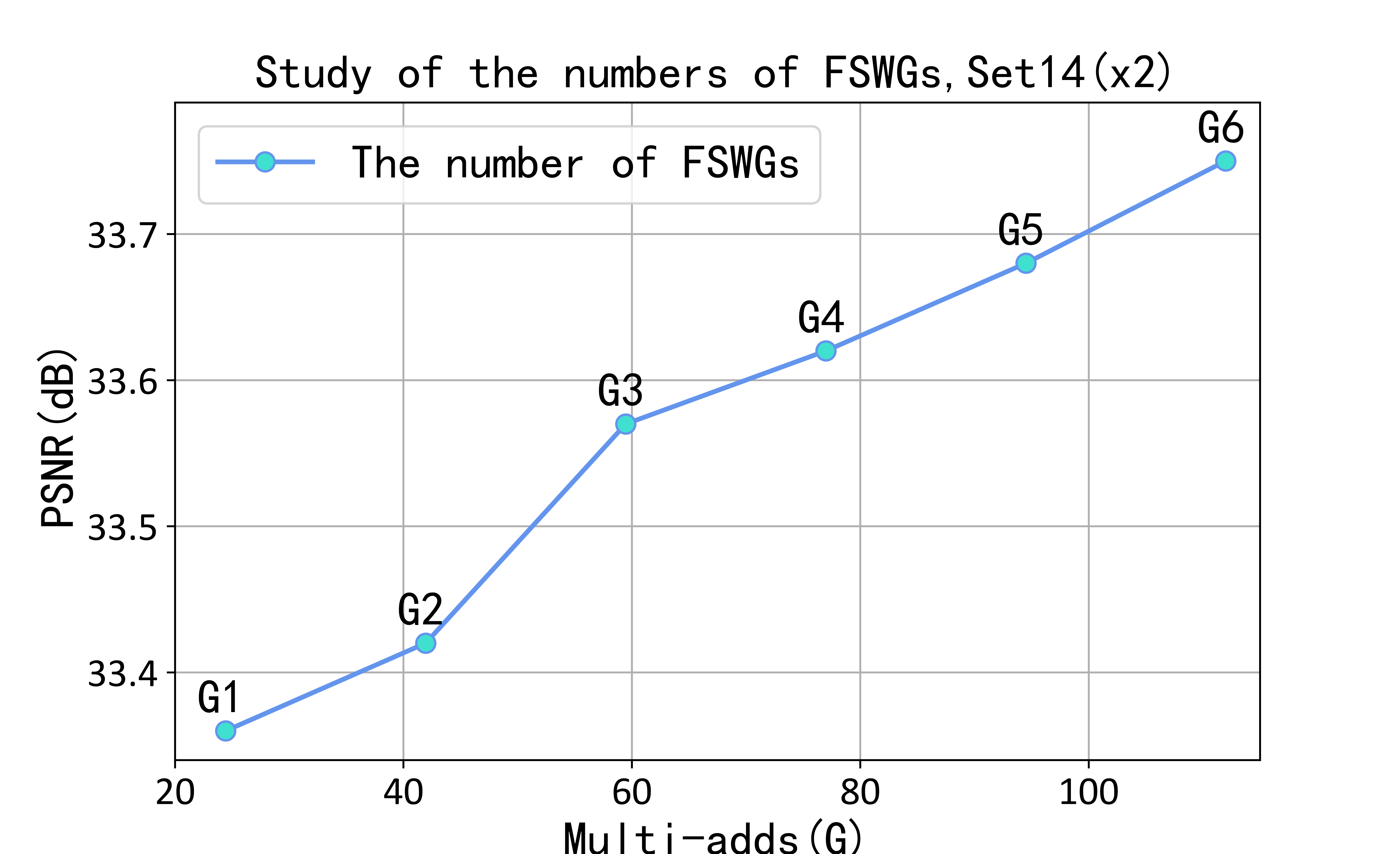}
   \\
   \includegraphics[width=8cm]{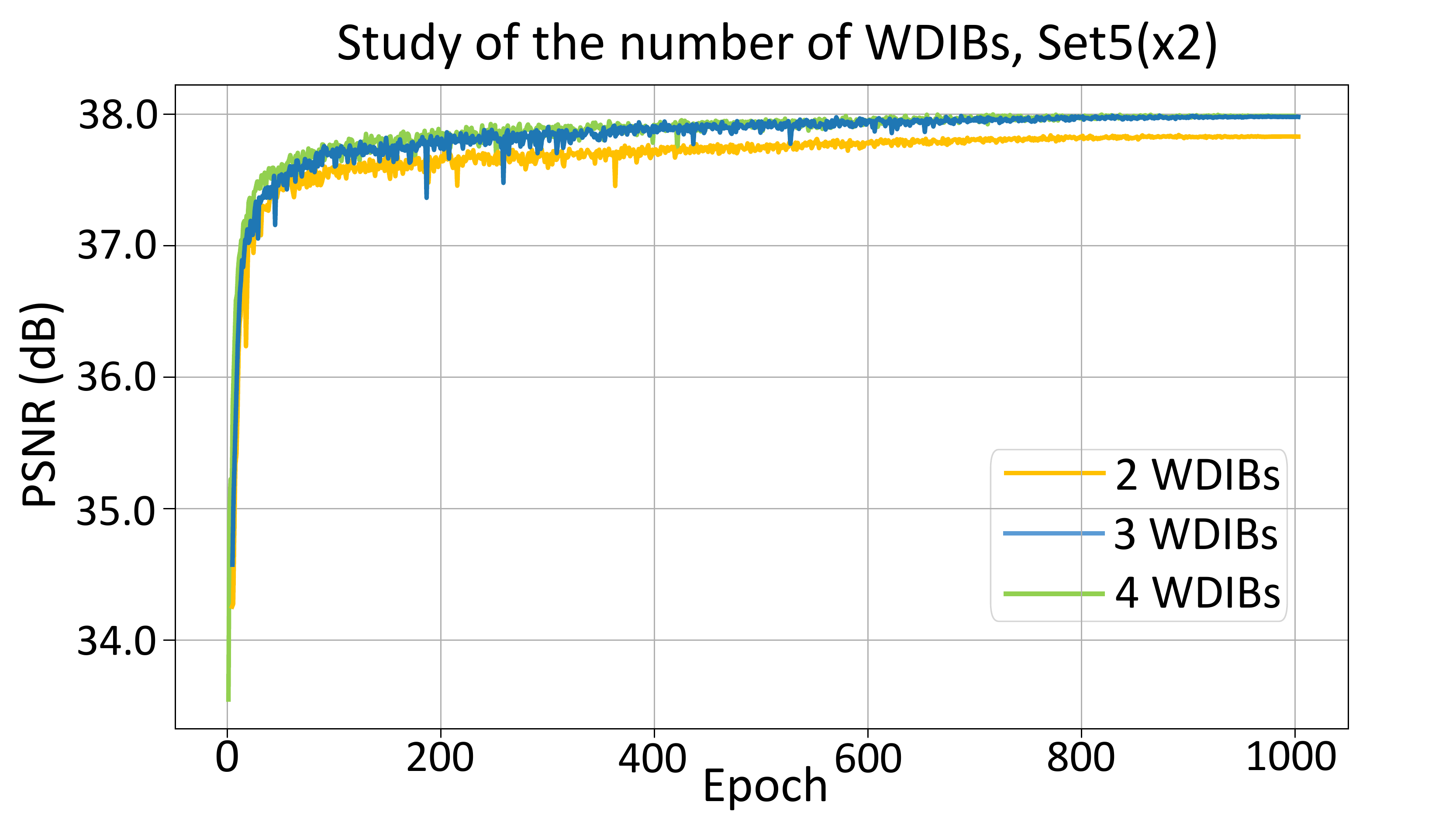}
   \caption{Study of different numbers of FSWGs and WDIBs.}
   \label{curve}
\end{figure}

\textbf{The combination structure of WDIB.} The FSWG in FDIWN is composed of three WDIBs, which are interacted with each other to yield more representative features. To verify the effect of this combination of WDIB, we chose three cascaded WDIBs as the baseline. In other words, three WDIBs are simply connected without any interaction. According to Table~\ref{tab3}, we can observe that the information blending structure we used is more effective. It improves the PSNR from 37.81 dB to 37.85 dB with limited computational costs, which further proves the effectiveness of the proposed information interaction structure. In addition, the long-skip connection provided by the WIRW unit can further improve model performance. All these experiments show that the combined structure of WDIB is effective.

\textbf{Efficiency trade-off.} In Figure~\ref{curve}, we show the performance change under different numbers of FWSGs and WDIBs. Among them, the circular point $Gm$ denotes that $m$ FSWGs are cascaded. According to the figure, we can observe that \romannumeral 1) As the number of FSWG and WDIB increases, the model performance can be further improved; \romannumeral 2) The PSNR does not increase when the number of WDIB increases from 3 to 4. Therefore, we use 6 FSWGs and 3 WDIBs in the final version model to achieve a good balance between model performance, size, and computational costs. 

\begin{figure}[t]
   \centering
   \includegraphics[width=8cm]{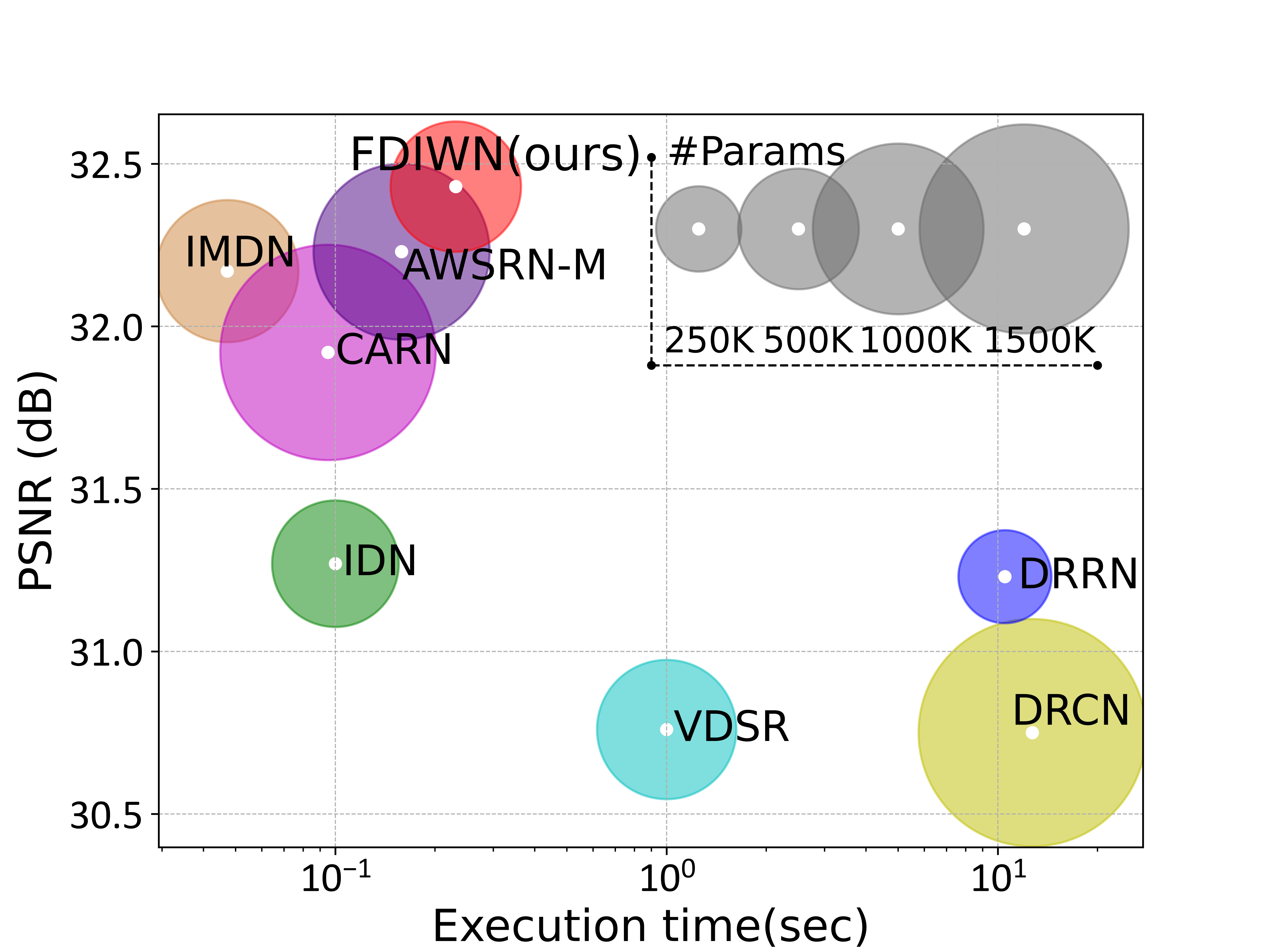}
   \caption{Inference speed study on Urban100 with x2 SR.}
   \label{time}
\end{figure}

\begin{figure}[h]
	\centerline{\includegraphics[width=9cm]{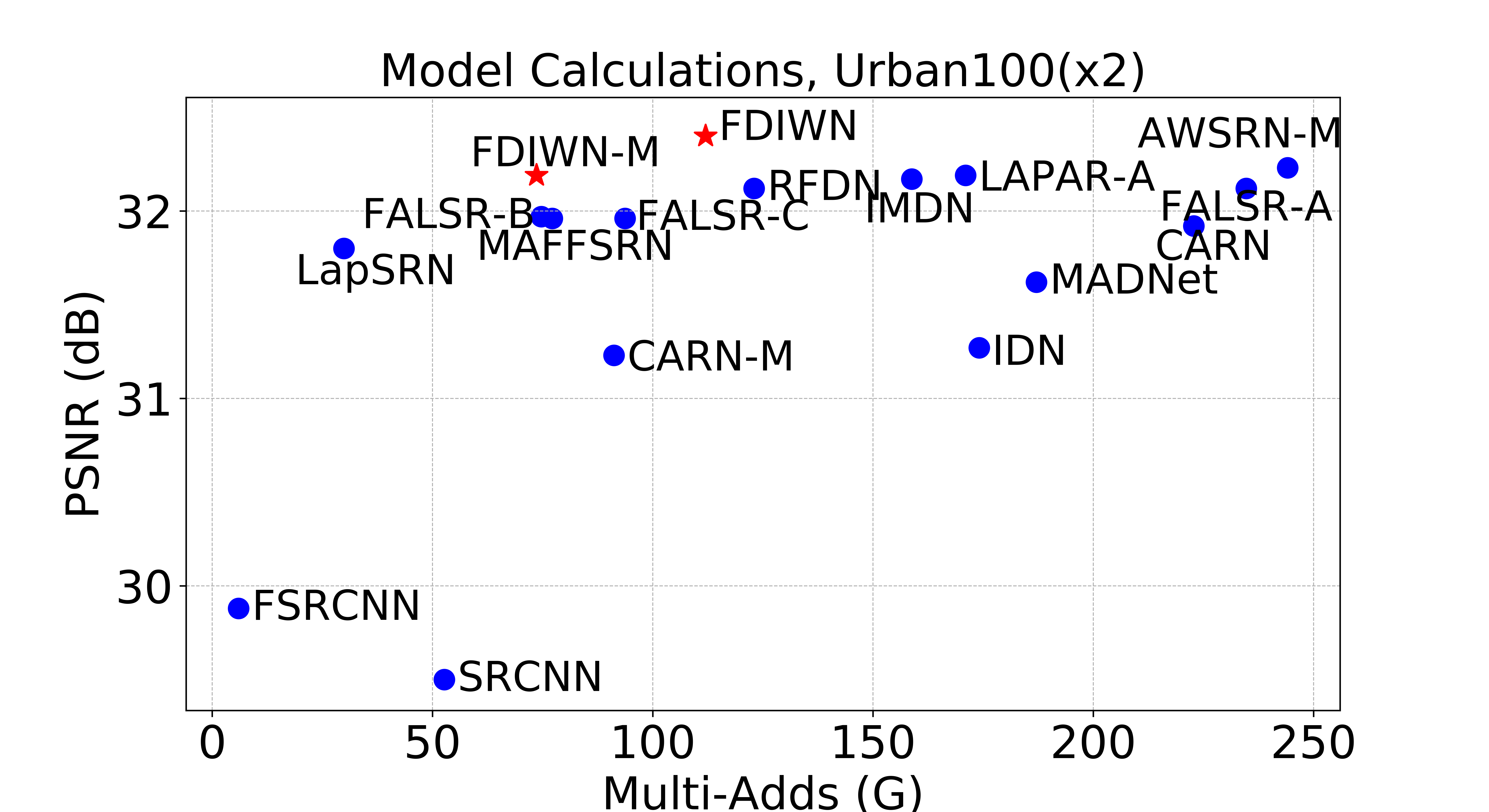}}
	\caption{Investigations of the model size and performance.}
	\label{size}
\end{figure}

\textbf{Model complexity analysis.} In Figure~\ref{time}, we show the execution time comparison with other classic lightweight SISR models. Obviously, FDIWN achieves competitive results with fewer parameters. Although our FDIWN is not the fastest model, the execution time is acceptable. Meanwhile, we show the multi-adds comparison between FDIWN and other SISR methods in Figure~\ref{size}. Obviously, our FDIWN achieves the best balance. Therefore, we can draw a conclusion that FDIWN gains a better trade-off between model size, performance, inference speed, and multi-adds.

\begin{table*}[t]
 \centering
  \setlength\tabcolsep{1.3mm}
  {\fontsize{9pt}{10pt}\selectfont
  \begin{tabular}{@{}l|c|c|c|cccc@{}}
   \toprule
    & &   &   & Set5  &Set14    & BSDS100  & Urban100 \\
   Algorithm           &Scale        &Params &Multi-adds & PSNR~~~SSIM     & PSNR~~~SSIM          & PSNR~~~SSIM    & PSNR~~~SSIM \\ \midrule                                                                                                                            
            %\hline
   SRCNN~\cite{3dong2015image}        &   & 57K & 52.7G  & 32.75~~~0.9090        & 29.30~~~0.8215            & 28.41~~~0.7863    & 26.24~~~0.7889     \\
   %FSRCNN~\cite{4dong2016accelerating}            &   & 12K  & 5.0G                   & 33.16~~~0.9140        &29.43~~~0.8242             & 28.53~~~0.7910    & 26.43~~~0.8080     \\
   VDSR~\cite{25kim2016accurate}      &   & 665K & 612.6G  & 33.67~~~0.9210        & 29.78~~~0.8320          & 28.83~~~0.7990    & 27.14~~~0.8290    \\
   DRCN~\cite{34kim2016deeply}          &   & 1774K &17974.3G  & 33.82~~~0.9226        & 29.76~~~0.8311            & 28.80~~~0.7963    & 27.15~~~0.8276    \\
   IDN~\cite{37hui2018fast}          &    & 590K &105.6G  & 34.11~~~0.9253        & 29.99~~~0.8354            & 28.95~~~0.8013    & 27.42~~~0.8359    \\
   CARN-M~\cite{1ahn2018fast}         &                      & 412K & 46.1G  & 33.99~~~0.9236        & 30.08~~~0.8367            & 28.91~~~0.8000    & 27.55~~~0.8385    \\
   CARN~\cite{1ahn2018fast}         &                      & 1592K & 118.8G  & 34.29~~~0.9255        & 30.29~~~0.8407            & 29.06~~~0.8034    & 28.06~~~0.8493    \\
   IMDN~\cite{11hui2019lightweight}         &$ \times 3$ & 703K  & 71.5G           & 34.36~~~0.9270        & 30.32~~~0.8417                            & 29.09~~~0.8046    & 28.17~~~0.8519    \\
   AWSRN-M~\cite{38wang2019lightweight}          &                      & 1143K & 116.6G  & 34.42~~~0.9275        & 30.32~~~0.8419   & 29.13~~~0.8059    & 28.26~~~0.8545    \\
   MADNet~\cite{35lan2020madnet}          &   & 930K & 88.4G           & 34.16~~~0.9253       & 30.21~~~0.8398                             & 28.98~~~0.8023    & 27.77~~~0.8439    \\
   RFDN~\cite{19liu2020residual}          &   & 541K & 55.4G           & 34.41~~~0.9273        & 30.34~~~0.8420                             & 29.09~~~0.8050    & 28.21~~~0.8525    \\
   MAFFSRN~\cite{40muqeet2020multi}                  &    & 418K  & 34.2G                & 34.32~~~0.9269        & 30.35~~~0.8429   & 29.09~~~0.8052    & 28.13~~~0.8521    \\
   LAPAR-A~\cite{36li2021lapar}                             &    & 594K & 114G           & 34.36~~~0.9267        & 30.34~~~0.8421                              & 29.11~~~0.8054   & 28.15~~~0.8523  \\
   \textbf{FDIWN-M (Ours)}       &   & 446K &35.9 G  & 34.46~~~0.9274     & 30.35~~~0.8423    & 29.10~~~0.8051    & 28.16~~~0.8528   \\
   \textbf{FDIWN (Ours) }      &   & 645K &51.5G  & \textbf{34.52}~~~\textbf{0.9281}   &\textbf{30.42}~~~\textbf{0.8438}  &\textbf{29.14}~~~\textbf{0.8065}   &\textbf{28.36}~~~\textbf{0.8567}                                                                                                       \\ 
   \hline
    SRCNN~\cite{3dong2015image}           &    & 57K & 52.7G  & 30.48~~~0.8628        & 27.49~~~0.7503                  & 26.90~~~0.7101     & 24.52~~~0.7221     \\
   %FSRCNN~\cite{4dong2016accelerating}            &    & 12K  & 4.6G                   & 30.71~~~0.8657        &27.59~~~0.7535                  & 26.98~~~0.7150    & 24.62~~~0.7280     \\
   VDSR~\cite{25kim2016accurate}      &    & 665K & 612.6G  & 31.35~~~0.8838        & 28.01~~~0.7674         & 27.29~~~0.7251    & 25.18~~~0.7524    \\
   DRCN~\cite{34kim2016deeply}          &    & 1774K &17974.3G  & 31.53~~~0.8854        & 28.02~~~0.7670            & 27.23~~~0.7233    & 25.14~~~0.7510    \\
   LapSRN~\cite{41lai2017deep}                   &    & 813K &149.4G                  & 31.54~~~0.8850        & 28.19~~~0.7720        & 27.32~~~0.7280    & 25.21~~~0.7560    \\
   IDN~\cite{37hui2018fast}                         &    & 590K &81.9G  & 31.82~~~0.8903        & 28.25~~~0.7730           & 27.41~~~0.7297    & 25.41~~~0.7632    \\
   CARN-M~\cite{1ahn2018fast}         &                      & 412K & 32.5G  & 31.92~~~0.8903        & 28.42~~~0.7762            & 27.44~~~0.7304    & 25.62~~~0.7694    \\
   CARN~\cite{1ahn2018fast}         &  & 1592K & 90.9G  & 32.13~~~0.8937        & 28.60~~~0.7806            & 27.58~~~0.7349    & 26.07~~~0.7837    \\
   IMDN~\cite{11hui2019lightweight}        &$ \times 4$    & 715K         & 40.9G           & 32.21~~~0.8948        & 28.58~~~0.7811                             & 27.56~~~0.7353   & 26.04~~~0.7838    \\
   AWSRN-M~\cite{38wang2019lightweight}          &                      & 1254K & 72.0G & 32.21~~~0.8954       & 28.65~~~\textbf{0.7832}    & 27.60~~~0.7368    & 26.15~~~0.7884    \\
   MADNet~\cite{35lan2020madnet}        &    & 1002K       & 54.1G           & 31.95~~~0.8917        & 28.44~~~0.7780                            & 27.47~~~0.7327    & 25.76~~~0.7746    \\
   RFDN~\cite{19liu2020residual}                  &    & 550K         & 31.6G           & \textbf{32.24}~~~0.8952        & 28.61~~~0.7819     & 27.57~~~0.7360    & 26.11~~~0.7858    \\
   MAFFSRN~\cite{40muqeet2020multi}                  &    & 441K  & 19.3G                & 32.18~~~0.8948        & 28.58~~~0.7812                            & 27.57~~~0.7361    & 26.04~~~0.7848    \\
   ECBSR~\cite{zhang2021edge}  &    & 603K  & 34.73G           & 31.92~~~0.8946        & 28.34~~~0.7817                            & 27.48~~~\textbf{0.7393}    & 25.81~~~0.7773    \\
   LAPAR-A~\cite{36li2021lapar}                             &    & 659K  & 94G                  & 32.15~~~0.8944       & 28.61~~~0.7818    & 27.61~~~0.7366    & 26.14~~~0.7871  \\
   \textbf{FDIWN-M (Ours)}              &    & 454K  & 19.6G  & 32.17~~~0.8941       & 28.55~~~0.7806        & 27.58~~~0.7364    &26.02~~~0.7844     \\
   \textbf{FDIWN (Ours) }              &    & 664K  &28.4G   & 32.23~~~\textbf{0.8955}   &\textbf{28.66}~~~0.7829  &\textbf{27.62}~~~0.7380   &\textbf{26.28}~~~\textbf{0.7919}                                                 \\ \bottomrule
  \end{tabular}}
   \caption{Quantitative comparison with state-of-the-art models. The best results are highlighted.}
  \label{tab4}
\end{table*}

\subsection{Comparisons with State-of-the-Art Methods}
\label{sec44}

In Table~\ref{tab4}, we provide the detail quantitative comparison with several representative SOTA SISR methods. According to the table, we can observe that \romannumeral 1) Models with a similar number of parameters to our model perform worse than ours; \romannumeral 2) The models with the same effects have more parameters than ours. Therefore, we can draw a conclusion that our proposed FDIWN-M and FDIWN stand out from these methods and perform very competitively in balancing model size, performance, and computational cost.

\begin{figure*}
	\centerline{\includegraphics[width=16cm, trim=0 50 0 0]{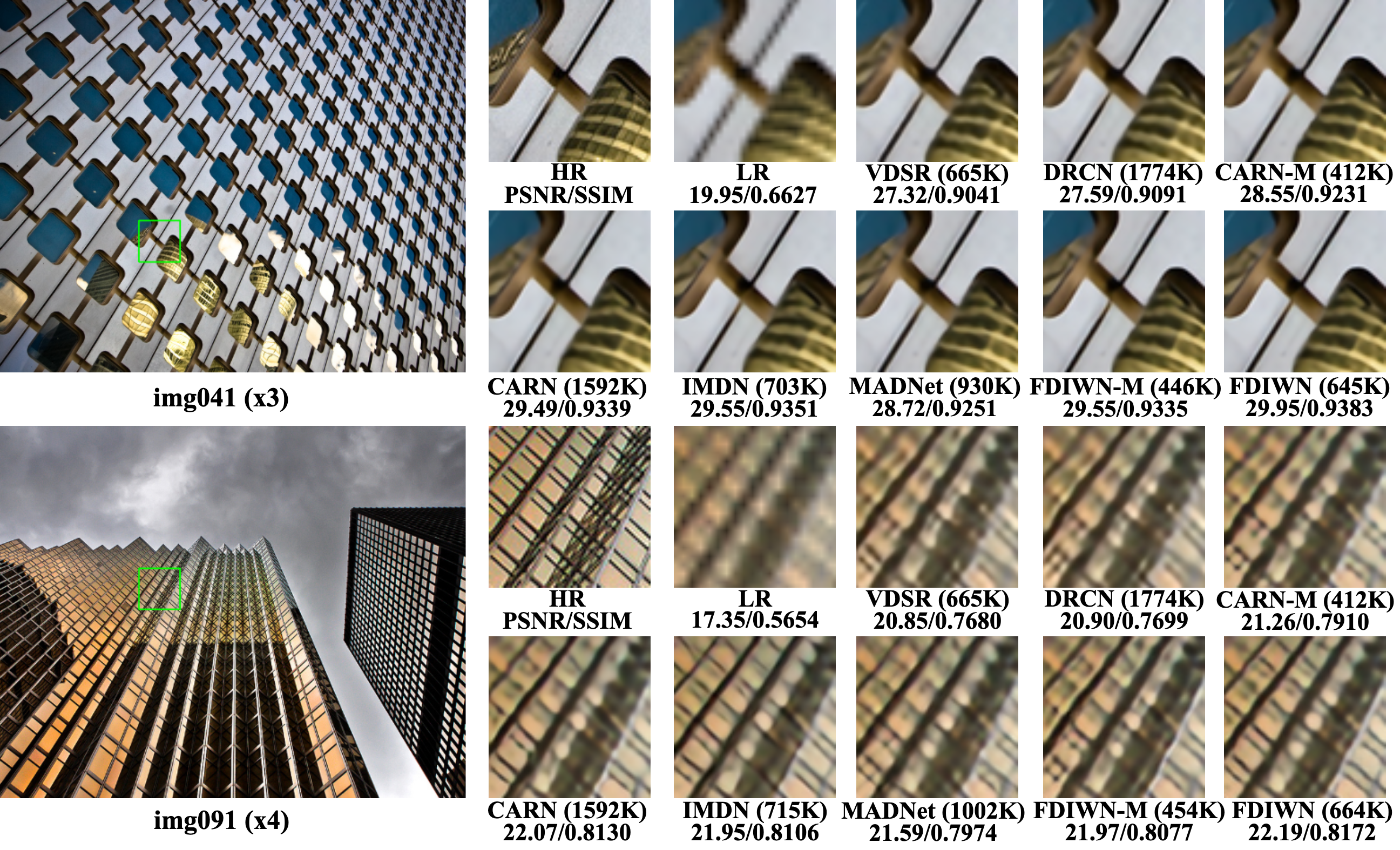}}
	\caption{Visual comparisons on the Urban100 dataset. Due to the page limit, please zoom in for details.}
	\label{Figure 9}
\end{figure*}

Figure~\ref{Figure 9} allows us to visually compare our method with other advanced methods on the Urban100 dataset. Through a horizontal comparison of the SR results, we can qualitatively see the effectiveness and excellence of our proposed method. Meanwhile, their PSNR and SSIM results are also provided. Our method not only has better visual details but also outperforms existing advanced methods in terms of quantitative data comparisons.

\section{Conclusions}

In this paper, we proposed an effective and lightweight Feature Distillation Interaction Weighting Network (FDIWN) for SISR. Compared to other lightweight SISR models, FDIWN not only reduces the computational overhead but also improves the SR performance. In summary, the improvement of our FDIWN is mainly due to the following parts: (\romannumeral 1) The specially designed wide-residual weighting units (including WIRW and WCRW) have a stronger ability to distill useful features than ordinary residual blocks; (\romannumeral 2) The shuffle attention (SA) mechanism makes the feature extraction concentrating on the key information; (\romannumeral 3) The well-designed wide-residual units based WRDC module and SCF module can flexibly aggregate and distill more representative features, allowing features from different scales to efficiently interact with each other. Therefore, the contextual and intermediate features can be well interacted, which benefits high-quality SR image reconstruction. Evaluation results on benchmarks have shown that the proposed FDIWN achieved a good balance between model size, performance, and computational cost.

\section{Acknowledgments}
This work was supported in part by the National Natural Science Foundation of China (Nos. 61972212, 61772568, 62076139 and 61833011), the Natural Science Foundation of Jiangsu Province (No. BK20190089), and the Six Talent Peaks Project in Jiangsu Province (No. RJFW-011).

% Use \bibliography{yourbibfile} instead or the References section will not appear in your paper
\bibliography{aaai22}

\end{document}